\documentclass[11pt]{article}

\usepackage{colacl}
\usepackage{graphicx}
\usepackage{pseudocode}
\usepackage{amsthm}

\newcommand*\rightlang{\stackrel{\rightarrow}{\mathcal L}}
\newcommand*\lang{\mathcal L}
\newcommand*\fstequiv{\equiv_{\mathrm{ST}}}

\newcommand*\myeqref[1]{\stackrel{(#1)}{=}}

\newcommand*\eqdef{\stackrel{\mathrm{def}}{=}}

\newcommand*\path[1]{\mathit{Path}_{#1}}

\newcommand*\collect{}

\newcommand*\bigdistrib{\bigwedge}

\newenvironment{abstractquote}{\begin{list}{}{\setlength{\rightmargin}{0.25in}\setlength{\leftmargin}{0.25in}}\item[]\small}{\end{list}}

\newtheorem{proposition}{Proposition}

\newtheorem{definition}{Definition}

\title{Incremental Construction of Minimal Acyclic Sequential Transducers
from Unsorted Data}
\author{
Wojciech Skut\\Rhetorical Systems Ltd\\4 Crichton's Close\\Edinburgh EH8 8DT, Scotland\\{\tt wojciech.skut@rhetorical.com}
}

\begin{document}
\maketitle
\bibliographystyle{acl}
\begin{abstract}
\begin{abstractquote}
This paper presents an efficient algorithm for the incremental
construction of a minimal acyclic sequential transducer (ST) for a
dictionary consisting of a list of input and output strings. The
algorithm generalises a known method of constructing minimal
finite-state automata \cite{Daciuk:ea:00}. Unlike the algorithm
published by Mihov and Maurel~\shortcite{Mihov:Maurel:01}, it does not
require the input strings to be sorted. The new method is illustrated
by an application to pronun\-ciation dictionaries.
\end{abstractquote}
\end{abstract}

\section{Introduction}
\label{sec:intro}
Sequential transducers constitute a powerful formalism for storing and
processing large dictionaries. Each word in the dictionary is
associated with an annotation, e.g., phonetic transcription or a
collection of syntactic features. Since STs are deterministic, lexical
lookup can be performed in linear time. Space efficiency can be
achieved by means of minimisation algorithms
\cite{Mohri:94,Eisner:03}.

In the present paper, we consider the following problem. Given a list
of strings $w^{(1)}\ldots w^{(m)}$ associated with annotations
$o^{(1)}\ldots o^{(m)}$, we want to construct a minimal ST $T$
implementing the mapping $f(w^{(j)}) = o^{(j)}, j=1\ldots m$.

The na{\"{\i}}ve way of doing that would be first to create a
(non-minimal) ST implementing $f$ and then to minimise it. As pointed out
by Daciuk et al.~\shortcite{Daciuk:ea:00}, this can be inefficient,
especially for large $m$. Instead, the same task can be performed more
efficiently in an {\em incremental} way, i.e., by constructing a
sequence of transducers $T_1\ldots T_m$ such that each $T_j$ is the
minimal ST implementing the restriction of the original mapping $f$ to
the first $j$ words ($f|_{\{w^{(1)}\ldots w^{(j)}\}}$). Since the insertion of a
new word $w^{j+1}$ typically affects only few states of the
transducer, $T_{j+1}$ can be constructed from $T_j$ by changing only a
small part of its structure.

Daciuk et al.~\shortcite{Daciuk:ea:00} show how to incrementally
construct a minimal finite-state automaton for a list of words
$w_1\ldots w_m$.  Their algorithm can also be applied to transducers,
but fails to produce a minimal ST in the general case. Mihov and
Maurel~\shortcite{Mihov:Maurel:01} describe an algorithm that handles
the ST case correctly, but requires the words to be sorted in
advance. In some applications, this requirement is unrealistic as
lexical entries may be added dynamically to an already constructed
dictionary.%
\footnote{Many systems employ a built-in lexicon that is constructed
off-line, but may be extended with user dictionaries merged in at
any time in any order.  The only way to use the sorted-data
algorithm would be to unfold the already minimised lexicon, add the new
entries, sort the data and re-apply the construction
method. \label{fn:whynot} } The present paper presents an algorithm
that does not make assumptions about the order of the list
$w^{(1)}\ldots w^{(m)}$.

The paper is structured as follows. Section~\ref{sec:def} introduces
definitions and notation. The ori\-ginal algorithm for finite automata
is described in section~\ref{sec:minfsa}. Section~\ref{sec:stprob}
explains why the algorithm does not work for transducers. The required
generalisation is introduced in section~\ref{sec:mincrit}; an
algorithm based on this generalisation is the topic of
section~\ref{sec:algo}. Section~\ref{sec:eval} illustrates the new
method with a practical application.

\section{Definitions}
\label{sec:def}

\begin{definition}{\bf(Deterministic FSA)}
A deterministic finite-state automaton (DFSA) over an alphabet
$\Sigma$ is a quintuple $A=(\Sigma, Q, q_0, \delta, F)$ such that $Q$
is a finite set of states, $q_0 \in Q$ is the initial state, $F
\subset Q$ the set of final states, and $\delta: Q \times \Sigma
\rightarrow Q$ is a (partial) transition function. $\delta$ can be
extended to the domain $Q \times \Sigma^*$ by the following
definition: $\delta^*(q,\epsilon) = q$, $\delta^*(q, wa) =
\delta(\delta^*(q,w),a)$.

$A$ accepts a string $w \in \Sigma^*$ if $q = \delta^*(q_0,w)$ is
defined and $q \in F$. The set of all strings accepted by $A$ is
called the language of $A$ and written ${\mathcal L}(A)$. An FSA is
called {\em trim} if every state belongs to a path from $q_0$ to a final
state.
\end{definition}

\begin{definition}{\bf (Right Language)}
Let $A = (\Sigma, Q, q_0, \delta, F)$ be a DFSA. The right language
$\rightlang_A(q)$ of a state $q$ in $A$ is the set of all strings $w$
such that $\delta^*(q,w) \in F$. If $\delta^*(q_0,u)$ is defined for $u \in
\Sigma^*$, the right language of $u$ is defined as $\rightlang_A(u)
\stackrel{\mathrm{def}}{=} \rightlang_A(\delta^*(q_0, u))$.  We omit the
subscript whenever $A$ can be inferred from the context.
\end{definition}

\begin{definition}{\bf (Sequential Transducers)}
A sequential transducer (ST) over an input alphabet $\Sigma$ and an
output alphabet $\Delta$ is a 7-tuple
$T=(\Sigma,\Delta,Q,q_0,\delta,\sigma,F)$ such that
$A=(\Sigma,Q,q_0,\delta,F)$ is a DFSA and $\sigma: Q \times \Sigma
\rightarrow \Delta^*$ is a function that labels transitions with
emissions from $\Delta^*$ ($\mathit{Dom}(\sigma)
=\mathit{Dom}(\delta)$).

Function $\sigma$ can be extended to $Q \times \Sigma^*$ according to
the following recursive definition: $\sigma^*(q,\epsilon) = \epsilon,
\sigma^*(q, wa) = \sigma^*(q,w)\collect \sigma(\delta^*(q,w),a)$.  

Unless indicated otherwise, the definitions formulated above for DFSAs
also apply to STs ($\lang(T)$, $\rightlang(u)$, $\rightlang(q)$).

Each ST $T$ realises a function $f_T: \Sigma^* \rightarrow \Delta^*$
such that $\mathit{Dom}(f_T) = \lang(T)$ and $f_T(u) = \sigma^*(q_0,u)$ for
$u \in \mathit{Dom}(f_T)$. A {\em
sequential} function is one that can be realised by an ST.
\end{definition}

\begin{definition}{\bf (Minimal FSA/ST)}
A DFSA $A=(\Sigma,Q,q_0,\delta,F)$ is called minimal if $|Q| \leq |Q'|$ for all
DFSA $A'=(\Sigma,Q',q_0',\delta',F')$ such that $\lang(A')=\lang(A)$. 

An ST $T=(\Sigma,\Delta,Q,q_0,\delta,\sigma,F)$ is called minimal if
$|Q| \leq |Q'|$ for any ST
$T'=(\Sigma,\Delta,Q',q_0',\delta',\sigma',F')$ such that $f_{T'}=f_{T}$.
\end{definition}

\subsection{String Notation}

For $u, v \in \Sigma^*$, $uv$ and $u\cdot v$ denote the concatenation
of $u$ and $v$. $u \wedge v$ is the longest common prefix of $u$ and
$v$. If $u$ is a prefix of $v$ (written $u \leq_p v$), $u^{-1}v$
denotes the remainder of $v$: $u \cdot u^{-1}v = v$. $|u|$ denotes the
length of $u$, while $u_1\ldots u_{|u|}$ are its letters. $u_{[j\ldots
k]}$ denotes the substring $u_j\ldots u_k$ of $u$ ($u_{[j\ldots
k]}=\epsilon$ if $j > k$). If $T$ is an ST, $w \wedge T$ stands for
the longest prefix $u\leq_p w$ such that $\delta^*(q_0,u)$ is defined.

\section{Minimisation of FSAs}
\label{sec:minfsa}

The algorithm decribed by Daciuk et al.~\shortcite{Daciuk:ea:00} is
iterative. In each iteration, given a minimal acyclic trim FSA
$A=(\Sigma,Q,q_0,\delta,F)$ and a word $w\in \Sigma^*$, the algorithm
creates a DFSA $A'=(\Sigma,Q',q_0,\delta',F')$ such that $A'$ is the
minimal automaton for the language $\lang(A) \cup \{w\}$.  $A'$ then
serves as input to the next iteration.

The key notion here is the {\em  equivalence of states}. Two states
$q_1,q_2\in Q$ are considered equivalent (written $q_1\equiv q_2$) iff
$\rightlang(q_1)=\rightlang(q_2)$. A well-known result states that a
trim FSA is minimal iff it does not contain a pair $q_1,q_2$ of
distinct but equivalent states:
\begin{equation}\label{eq:fsaequiv}
\forall_{q_1,q_2\in Q}: q_1 \neq q_2 \Rightarrow q_1\not\equiv q_2
\end{equation}
Each iteration consists of two steps, which can be called {\em
insertion} and {\em local minimisation}.%
\footnote{The following description refers to a simpler variant of the
algorithm rather than to the optimised version described in the
pseudocode in the original publication. Optimisation is discussed
separately in section~\ref{sec:compl}.}

\subsection{Insertion}

The insertion operation identifies the longest prefix $w_1\ldots w_l$
of $w$ in $A$ and the corresponding states $q_0\ldots q_l$. Some of
them may be {\em confluence states}, i.e., states $q_i$ such that
$\mathit{in-degree}(q_i) > 1$. In order to prevent overgeneration, the algorithm
identifies the first confluence state $q_k$ and clones the path
$q_k\ldots q_l$. The cloned states $\hat{q}_k\ldots\hat{q}_l$ are
copies of the original ones, i.e., $\hat{\delta}(q_i,a) =
\delta(q_i,a)$ for all $a\in\Sigma$ such that $\delta(q_i,a)$ is
defined -- with the only exception of the transition consuming the
next symbol of $w$: $\hat{\delta}(\hat{q}_i,w_{i+1})=\hat{q}_{i+1}$.
Furthermore, $\hat{\delta}(q_{k-1},w_k) := \hat{q}_k$.

In the next step, a chain of states $\hat{q}_{l+1}\ldots \hat{q}_t$,
$\hat{q}_t\in\hat{F}$, consuming the remainder of $w$ (i.e.,
$w_{l+1}\ldots w_t$) is appended to $\hat{q}_l$ (if it has been
created) or to $q_l$. If $l=t$, the remainder is the empty string, and
we make sure that $q_l/\hat{q}_l \in \hat{F}$. 

Formally, this step creates an automaton 
$\hat{A}=(\Sigma,\hat{Q},q_0,\hat{\delta},\hat{F})$ such that, for
$i \in \{k,\ldots,t\}$:%
\footnote{We set $k := l+1$ if there are no confluence states.}
\begin{eqnarray*}
\hat{Q} & = & Q \cup \{\hat{q}_k\ldots \hat{q}_t\}\\
\hat{F} & = & F \cup \{\hat{q}_i: q_i \in F\}\\
\hat{\delta}(q,a) & = & \left\{ \begin{array}{l@{\ :\ }l}
                        \hat{q}_k & q=q_{k-1}, a=w_k\\
                        \hat{q}_{i+1} &  q=\hat{q}_i,a=w_{i+1}\\
                         \delta(q,a)  & \mbox{ otherwise. }
\end{array} \right.
\end{eqnarray*}
This completes the insertion step. The new automaton $\hat{A}$
obviously accepts the language $\lang(A) \cup \{w\}$ and preserves the
right languages of all states except $q_0\ldots q_{k-1}$.

\subsection{Local Minimisation}

The situation after insertion is that $\hat{A}$ contains
\begin{itemize}
\item a path $q_0\ldots q_{k-1}$ of states whose right languages may
have changed, 

\item a path $\hat{q}_k\ldots \hat{q}_t$ of newly created (partly
cloned) states, and

\item the remaining states of $A$, whose right languages have not
changed (i.e., (\ref{eq:fsaequiv}) still holds for $Q \backslash
\{q_0\ldots q_{k-1}\}$).
\end{itemize}
In order to make the new FSA minimal, the algorithm must enforce
condition~(\ref{eq:fsaequiv}) for the states $q_0\ldots
q_{k-1}\hat{q}_k\ldots \hat{q}_t$ by replacing them, if possible, by
their equivalents in a set $Q_{\not\equiv}$, which is initially set to
$Q \backslash \{q_0\ldots q_{k-1}\}$.

The sequence is traversed in reverse order, starting from
$\hat{q}_t$. In the $j$-th iteration ($j=1\ldots t-2$), the algorithm
checks if there is already a state $q' \in Q_{\not\equiv}$ equivalent
to the current state $q$. If such a $q'$ exists, $q$ is replaced by
$q'$ (i.e., $q$ is deleted and all transitions reaching $q$ are
redirected to $q'$). Otherwise, $q$ is added to $Q_{\not\equiv}$. In
this way, the algorithm gets rid of duplicates w.r.t. the equivalence
relation $\equiv$.  The automaton left after the last iteration
satisfies condition~(\ref{eq:fsaequiv}), i.e., it is minimal.

\subsection{The State Register}

The efficiency of the algorithm depends on its ability to quickly
check the equivalence of states. This check is fast because
$\delta^*(q, u) \in Q_{\not\equiv}$ for all $q\in Q_{\not\equiv}$ (at
any stage of the local minimisation step). In effect, $q_1 \equiv q_2
\iff Out(q_1) = Out(q_2) \wedge (q_1,q_2 \in F \vee q_1,q_2 \not\in
F)$, where $Out(q) = \{ (a,q'): \delta(q,a) = q'\}$ is the set of
transitions leaving $q$. Thus, for each $q\in Q_{\not\equiv}$,
the pair $(Out(q),q)$ is put on a {\em register}, i.e., an
associative container that maps sets of pairs
$(\mathit{input},\mathit{state})$ (uniquely identifying a right
language) to the corresponding states in $Q_{\not\equiv}$.

\section{Application to Transducers}
\label{sec:stprob}

\begin{figure}[h]
\begin{center}
\includegraphics[width=.4\linewidth]{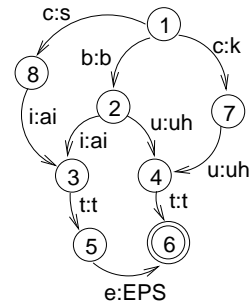}
\end{center}
\caption{A ``phonologically motivated'' alignment
of input and output symbols.}\label{fig:nondet}
\end{figure}

The problem for sequential transducers can be stated as follows: given
a minimal ST $T$ implementing a sequential function $f$, we want to
insert into $T$ a string $w$ associated with an emission $o$, creating
a minimal ST for $f \cup \{(w,o)\}$.  Daciuk et
al.~\shortcite{Daciuk:ea:00} state on this topic:
\begin{quotation}
This new algorithm can also be used to construct transducers. The
alphabet of the (transducing) automaton would be $\Sigma_1\times
\Sigma_2$, where $\Sigma_1$ and $\Sigma_2$ are the alphabets of the
levels. Alternatively, as previously described, elements of
$\Sigma_2^*$ can be associated with the final states of the dictionary
and only output once a valid word from $\Sigma_1^*$ is recognised.
\end{quotation}
Unfortunately, both suggested solutions are problematic. They require
that we commit ourselves to a particular alignment of input and output
symbols in the transitions in advance, before running the
algorithm. For instance, consider the fragment of a pronunciation
dictionary shown below.
{\tt\small
\begin{verbatim}
but | b uh t
bite | b ai t
cut | k uh t
cite | s ai t
\end{verbatim}
} Obviously, there are several string-to-string transducers that
implement this dictionary. One possibility would be to encode the
mapping in a phonologically motivated way, i.e., associating each
phonetic symbol with the grapheme(s) it corresponds to. Unfortunately,
the result of applying an FSA minimisation algorithm is
non-deterministic (figure~\ref{fig:nondet}).

The second suggestion made by Daciuk et al.~\shortcite{Daciuk:ea:00},
the use of final emissions, can be emulated using a special
end-of-string symbol \$. The result of FSA minimisation, shown at the
top of figure~\ref{fig:minimal}, is an ST, but not minimal, since it
has more states than the ST shown below.

\begin{figure}[h]
\begin{center}
\includegraphics[width=.60\linewidth]{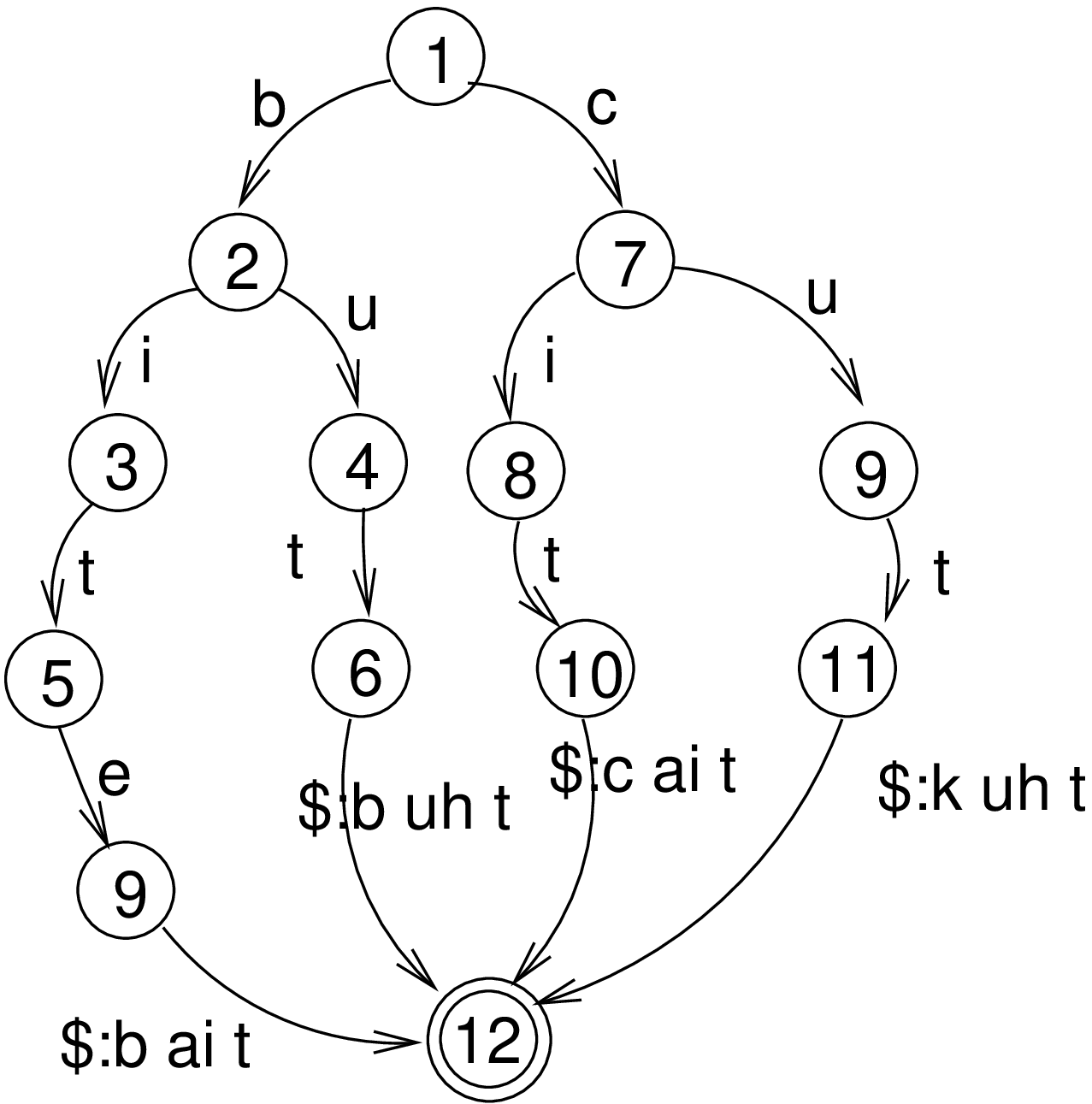}
\includegraphics[width=.52\linewidth]{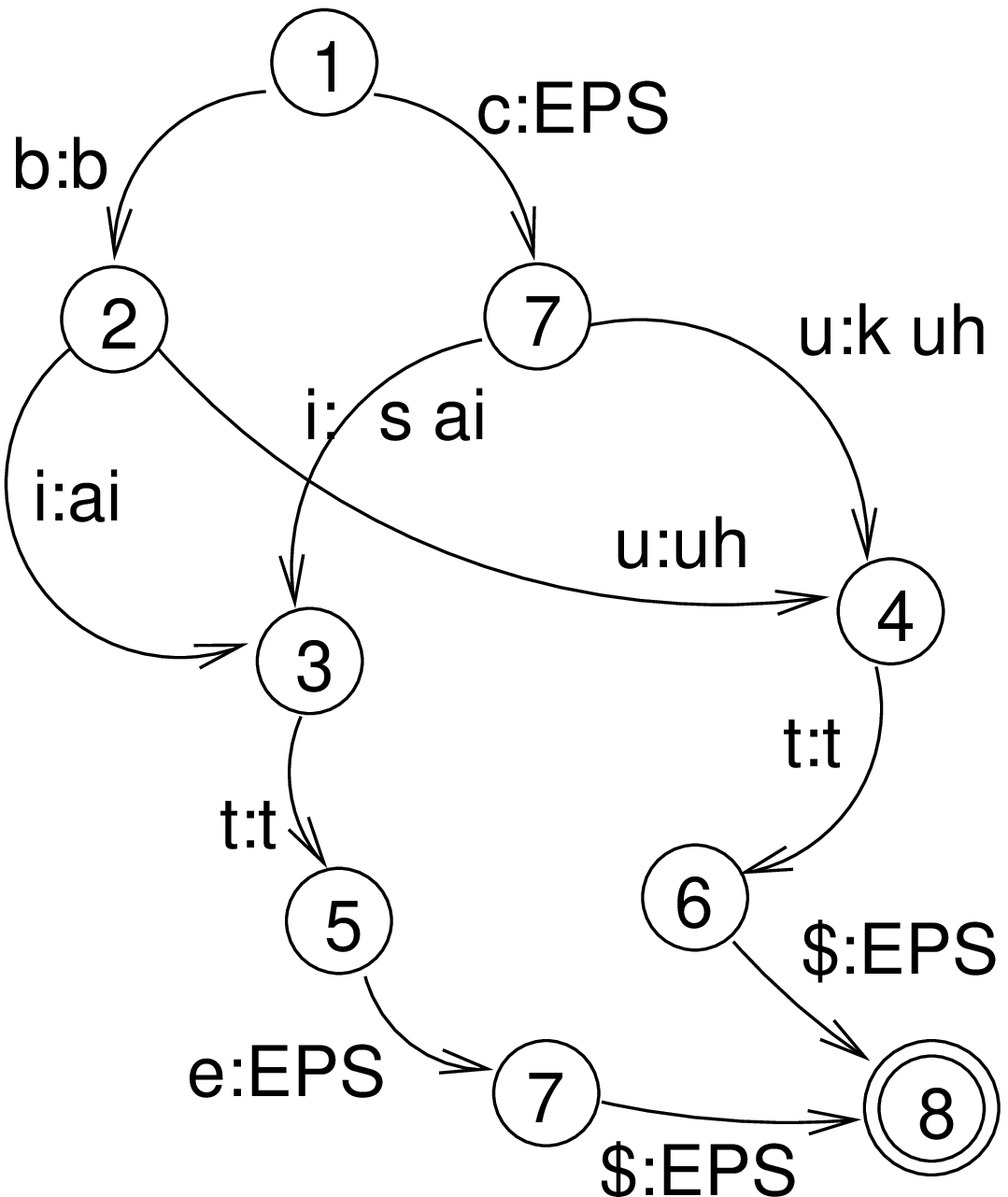}
\end{center}
\caption{An ST with final emissions and a mi\-nimal ST.}\label{fig:minimal}
\end{figure}

\section{ST Minimality Criteria}
\label{sec:mincrit}
In order to adapt the original algorithm to transducers, we need to
find an ST counterpart of the $\equiv$ relation defined for
finite-state automata. The following proposition constitutes a good
point of departure.

\begin{proposition}\label{prop:minfst}
\cite{Mohri:94} If $f:\Sigma^*\rightarrow \Delta^*$ is a sequential
function, there exists a minimal FST
$T=(\Sigma,\Delta,Q,i,\delta,\sigma,F)$ realising $f$. The size $|T|$
($=|Q|$) of $T$ is equal to the count of the equivalence relation
$R_f$ defined as
\begin{eqnarray}
u R_f v & \iff & \rightlang(u) =\rightlang(v)\ \wedge \label{eq:defrf}
\\
& \exists_{u',v' \in \Delta^*}      & \forall_{w \in \rightlang(u)} u'^{-1}f(uw) = v'^{-1}f(vw)  \nonumber  
\end{eqnarray}
\end{proposition}

$R_f$ is defined on the set $(\Sigma^*)^2$. In order to adapt the
original algorithm, we must define an equivalence relation on $Q$,
analogous to $\equiv$. It turns out that this is possible for
transducers that are {\em prefix-normalised}, as stated in the
following definition.
\begin{definition}\label{def:prefixnorm}
A sequential transducer $T=(\Sigma,\Delta,Q,i,\delta,\sigma,F)$ is
prefix-normalised if
\begin{equation}
\forall_{u \in \Sigma^*, \rightlang(u) \neq \emptyset} \sigma(q_0,u) = \bigdistrib_{z \in  \rightlang(u)} f_T(uz)
\label{eq:normalizedfst}
\end{equation}
\end{definition}
The following proposition allows us to define an equivalence relation
on $Q$ analogous to $R_f$.
\begin{proposition}\label{prop:equivclass}
If a trim sequential transducer $T=(\Sigma,\Delta,Q,i,\delta,\sigma,F)$ that
realises a function $f:\Sigma^*\rightarrow \Delta^*$ is
prefix-normalised, then the count of $R_f$ is equal to the count of
the relation $\fstequiv\subset Q^2$ defined as follows:
\begin{eqnarray*}
q \fstequiv q' & \iff & \rightlang(q) = \rightlang(q')\wedge  \\
	& & \forall_{w\in\rightlang(q)} \sigma^*(q,w) = \sigma^*(q',w)
\end{eqnarray*}
\end{proposition}
\begin{proof}
Since $T$ is trim, it is sufficient to prove
\[u R_f v \iff \delta^*(q_0,u) \fstequiv \delta^*(q_0,v)\]
\noindent
$\mathbf{\Leftarrow}$: follows immediately with $u'=\sigma^*(q_0,u)$,
$v'=\sigma^*(q_0,v)$, since $\rightlang(u)
\eqdef\rightlang(\delta^*(q_0,u))$.

\noindent
$\mathbf{\Rightarrow}$: Let $q :=
\delta^*(q_0,u)$, $q' := \delta^*(q_0,v)$. If $u R_f v$ then
$\rightlang(q)\myeqref{\ref{eq:defrf}}\rightlang(q')$
and there exist $u', v' \in \Delta$ such that: 

\[\forall_{w\in \rightlang(q)}: u'^{-1}f(uw) = v'^{-1}f(vw)\]

Since $f(uw)=\sigma^*(q_0,u)\collect\sigma^*(q,w)$, this is equivalent to:

\[u'^{-1}\sigma^*(q_0,u)\collect\sigma^*(q,w)=v'^{-1}\sigma^*(q_0,v)\collect\sigma^*(q',w)\]

Furthermore, $u'$ and $v'$ must be prefixes of $\sigma^*(q_0,u)$ and
$\sigma^*(q_0,v)$, respectively (otherwise, $T$ would not be
prefix-normalised), thus there exist $u'', v''$ such that
$u'\collect u''=\sigma^*(q_0,u), v'\collect v''=\sigma^*(q_0,v)$ and

\[\forall_{w\in \rightlang(q)}:u''\sigma^*(q,w)  =  v''\sigma^*(q',w)\]

This holds only if $u''$ is a prefix of $v''$ or vice versa. Without
loss of generality assume $v''=u''\collect z$. Then it follows:

\[\forall_{w\in\rightlang(u)} \sigma^*(q,w)  =  z\sigma^*(q',w)\]

Therefore, for all $w\in\rightlang(q')$, $z$ is a prefix of
$\sigma^*(\delta^*(q_0,v),w)$. Since $T$ is prefix-normalised, this
implies $z=\epsilon$, i.e.  $u''=v''$, hence
$\forall_{w\in\rightlang(u)} \sigma^*(\delta^*(q_0,u),w) =
\sigma^*(\delta^*(q_0,v),w)$, i.e., $q \fstequiv q'$.
\end{proof}

\section{The Algorithm}
\label{sec:algo}

According to proposition~\ref{prop:equivclass}, a modification of the
original algorithm (by Daciuk~et~al.~\shortcite{Daciuk:ea:00}) shall
produce a minimal ST if $\equiv$ is replaced by $\fstequiv$, and the
transducer being constructed is prefix-normalised in each
iteration. As in the original approach, each iteration is a two-step
operation: first, a new word-output pair $(w,o)$ is inserted into a
minimal, trim and prefix-normalised transducer $T$, creating a
prefix-normalised, ``almost minimal'' transducer $\hat{T}$. In the
second step, the ``redundant'' states on the path of $w$ are merged
with equivalent states in $\hat{T}$, resulting in a minimal, trim
and prefix-normalised ST implementing $f_T \cup \{(w,o)\}$.

The modifications to the FSA algorithm required in order to adapt it
to sequential transducers are discussed in
sections~\ref{sec:fstinsert} and \ref{sec:fstlocmin}. The pseudocode
is presented in section~\ref{sec:pseudo}.

\subsection{Insertion}
\label{sec:fstinsert}

Like the original algorithm, the modified one identifies $w\wedge T =
q_0\ldots q_l$ and creates new states $\hat{q}_{l+1}\ldots \hat{q}_t$.
It also clones the path $q_k\ldots q_l$ from the first confluence
state (if there is one).

The main complication is due to the insertion of the output sequence
$o$. In order for the ST to be well-formed, the output
$\hat{\sigma}^*(q_0,w\wedge T)$ generated by the prefix $w\wedge T$
must be a prefix of $o$. However, the original output
$\sigma^*(q_0,w\wedge T)$ in $T$ might not meet this requirement.

The solution is to ``push'' some of the outputs away from the path of
the prefix $w\wedge T = w_{[1\ldots l]}$. However, one must be careful
not to change the right languages (and their translations) of the
states that are not on the path of $w$. Furthermore,
proposition~\ref{prop:equivclass} requires the resulting ST to be
prefix-normalised.

\begin{figure}[h]
\begin{center}
\includegraphics[width=.9\linewidth]{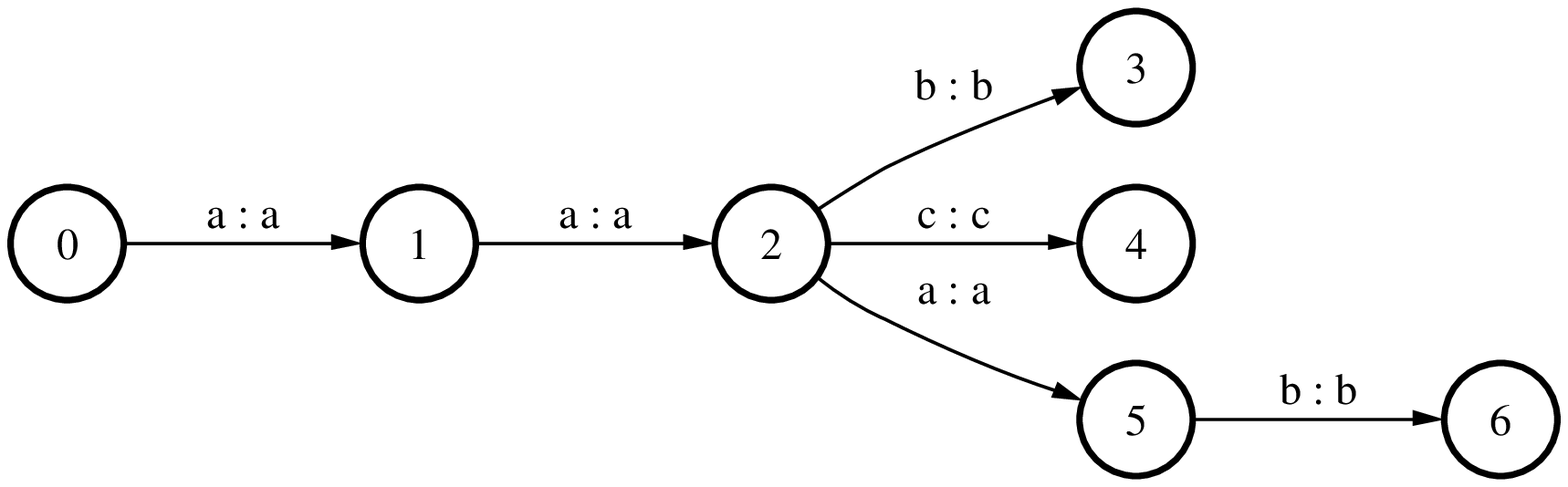}

\includegraphics[width=.9\linewidth]{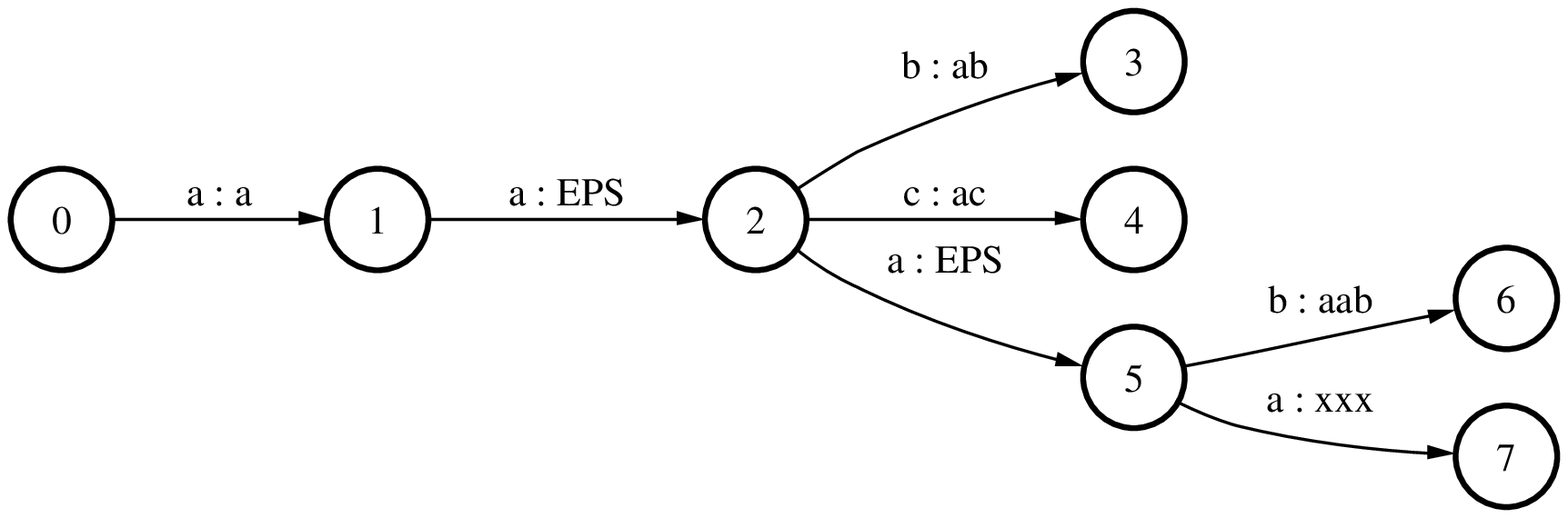}
\end{center}
\caption{Insertion of pair $(aaaa,axxx)$.}\label{fig:insert}
\end{figure}

All this can be achieved by the following recursive definition of
$\hat{\sigma}$.%
\footnote{The symbol $r_i$ ranges over $q_i, \hat{q}_i$ (whenever the
latter is defined). If there is no confluence state, $k :=
l+1$.}
\begin{eqnarray*}
\hat{\sigma}(r_i, w_{i+1}) & := & (\hat{\sigma}^*(q_0,w_{[1\ldots i]}))^{-1} \\
                           &   & \quad\quad    (o \wedge \sigma^*(q_0, w_{[1\ldots i+1]}))\\
\hat{\sigma}(r_i, a)       & := & (\hat{\sigma}^*(q_0,w_{[1\ldots i]}))^{-1}\\
	                   &    & \quad\quad \sigma^*(q_0, w_{[1\ldots i]}a), a \neq w_{i+1}
\end{eqnarray*}
An example of insertion is shown in figure~\ref{fig:insert}.

As for the suffix $w_{[l+1\ldots t]}$, we associate the output
remainder $\hat{\sigma}^*(q_0, w_{[1\ldots l]})^{-1}o$ with the
first transition. The remaining ones emit $\epsilon$.

This insertion mechanism ensures that $\hat{T}$ implements
the mapping $f_{T} \cup \{ (w,o)\}$ without changing the equivalence
of states not on the path $q_0\ldots q_{k-1}$.

\subsection{Local Minimisation}
\label{sec:fstlocmin}

As in the FSA algorithm, the path $q_0\ldots q_{k-1}\hat{q}_k\ldots
\hat{q}_t$ is traversed in reverse order. Each state $q$ for which
there exists an equivalent state $q'\in Q_{\not\equiv}$ is replaced by
$q'$.

What changes is the way the equivalence of two states $q_1,q_2$,
$q_2\in Q_{\not\equiv}$ is determined. Obviously, now it is no longer
sufficient to compare the target states and input labels of all
transitions leaving $q_1$ and $q_2$. However, since $\hat{T}$ is
prefix-normalised, the only extra bit to check is the output, i.e., we
keep the original formula $q_1 \fstequiv q_2 \iff Out(q) = Out(q')
\wedge (q_1,q_2 \in F \vee q_1,q_2 \not\in F)$, but redefine $Out(q)$
as $\{ (a,o,q') : \delta(q,a) = q' \wedge \sigma(q,a)= o \}$.


\subsection{Pseudocode}
\label{sec:pseudo}

In each iteration of the main loop, the algorithm takes a pair
$(\mathit{Word},\mathit{Output})$ and calls the procedures {\sc
insert()} and {\sc remove\_duplicates()}, responsible respectively for
the insertion and the local minimisation step. The former traverses
the word from left to right, clones the path from the first confluence
state down (if there is any), and ends up in the state $\delta^*(q_0,
w\wedge T)$, or its copy.  From there, {\sc insert\_suffix()} is
called, creating a chain of states corresponding to the remainder of
$\mathit{Word}$ (The pseudocode of {\sc insert\_suffix()} is omitted
for space reasons, but its functionality is very simple: the remainder
of $\mathit{Output}$ is emitted in the first transition of this new
chain of states).%
\footnote{If $w\wedge T = w$ and there is some output left,
then $f_{T} \cup \{(w,o)\}$ is not sequential.}
In each iteration of the {\tt for}-loop, the variable $\mathit{Output}$
holds the remainder of the original output that has not been emitted
so far. {\sc push\_outputs($\mathit{State,Residual}$)} takes care of
making the path outputs in $\hat{T}$ compatible with
$o=f_{\hat{T}}(o)$. After $i$ iterations of the loop, the argument
$Residual$ holds the value of $(o \wedge \sigma^*(q_0,w_{[1\ldots
{i-1}]}))^{-1}\sigma^*(q_0,w_{[1\ldots i]})$, i.e., the remainder of
$\sigma^*(q_0,w_{[1\ldots {i-1}]})$ after subtracting the longest
common prefix with $o$. This prefix is prepended to the output labels
of all transitions starting in $\mathit{State}$.%
\footnote{Note
that if $\mathit{State}\in F$ and $\mathit{Output}\neq \epsilon$,
$f_{T}\cup \{(w,o)\}$ is not sequential.}

The procedure {\sc remove\_duplicates()} traverses the path of $w$ in
$\hat{T}$ (in reverse order) and removes those states for which there
is an equivalent state in the register. The remaining states are added
to the register (note that the procedure {\sc insert\_suffix()}
de-registers all states from the root down to the first confluence state
-- if such a state exists -- or to the end of $w\wedge T$).

{\small\begin{pseudocode}[plain]{ConstructMinST}{ } \MAIN
   \mathit{Register} \GETS \emptyset\\ \WHILE \mbox{there is a
   word-output pair} \DO \BEGIN \mathit{(Word,Output)} \GETS \mbox{
   next pair }\\ \CALL{insert}{\mathit{Word, Output}}\\ \CALL
   {remove\_duplicates}{\mathit{Word}}\\ \END \\ \ENDMAIN\\ \\
   \PROCEDURE{insert}{\mathit{Word, Output}} \mathit{State} \GETS
   q_0\\ \mathit{FoundConfluence} \GETS \mathit{false}\\

\FOR i \GETS 1 \TO \mathit{ size(Word)} \DO
\BEGIN 
    \IF \mathit{State} \in  \mathit{Register}
    \THEN \mathit{Register} \GETS \mathit{Register} \backslash \{\mathit{State}\}\\
    \mathit{Symbol} \GETS \mathit{Word}[i]\\
    \mathit{Child} \GETS \delta(\mathit{State}, Symbol)\\ 
        \IF \mathit{Child} = \emptyset \mbox{  {\bf break}}
   \\

    \IF \mathit{InDegree(Child)}>1
            \THEN \mathit{FoundConfluence} \GETS \mathit{true}\\

    \IF \mathit{FoundConfluence}
         \THEN \delta(\mathit{State}, \mathit{Symbol}) \GETS \CALL{clone}{\mathit{Child}}\\

    \mathit{OutputPrefix} \GETS \mathit{Output} \wedge \sigma(\mathit{State,Symbol})\\
    \mathit{OutputSuffix}  \GETS \\
	\quad\quad\quad\mathit{OutputPrefix}^{-1}\sigma(\mathit{State,Symbol})\\
    \mathit{Output} \GETS \mathit{OutputPrefix}^{-1}\mathit{Output}\\
    \sigma(\mathit{State,Symbol}) \GETS \mathit{OutputPrefix}\\
    
    \mathit{State} \GETS \delta(\mathit{State}, \mathit{Symbol})\\
    \CALL{push\_outputs}{\mathit{State,OutputSuffix}}\\
\END\\
\ROF \\
\CALL{insert\_suffix}{\mathit{State},\mathit{Word}_{[i..\mathit{size(Word)}]},\mathit{Output}}
\ENDPROCEDURE
\PROCEDURE{remove\_duplicates}{\mathit{Word}}
\FOR i \GETS \mathit{size(Word)} - 1 \DOWNTO 1\\
\BEGIN
   \mathit{State} \GETS \delta^*(q_0, \mathit{Word}[1\ldots i])\\
   \mathit{Symbol} \GETS \mathit{Word}[i+1]\\
   \mathit{Child} \GETS \delta(\mathit{State,Symbol})\\
   \IF \exists q\in \mathit{Register}, q \fstequiv \mathit{Child} 
  \THEN \delta(\mathit{State,Symbol}) \GETS q
   \ELSE \mathit{Register} \GETS \mathit{Register} \cup \{\mathit{Child} \}
\END\\
\ROF
\ENDPROCEDURE
\PROCEDURE{push\_outputs}{\mathit{State,Residual}}
 \FOREACH a \in \Sigma\\
 \BEGIN
      \IF \delta(\mathit{State},a) \mbox{ is defined}
        \THEN     \sigma(\mathit{State},a) \GETS Residual \cdot \sigma(\mathit{State},a)
 \END
 \ENDPROCEDURE
\end{pseudocode}
}

\vspace{-1.2cm}

\subsection{Extensions}
\label{sec:ext}

The algorithm can be extended to the more general case of {\em
subsequential transducers} (SST). An SST is an ST that emits {\em
final outputs} when halting in a final state \cite{Mohri:97}. It can
be emulated in the ST framework by appending a special end-of-string
character $\$$ to each string, making each final state $q$ in the
transducer non-final and adding a transition from $q$ via $\$$ to a
new final state $q_f$. The output associated with this transition is
the ST equivalent of a final output. In this encoding, the new
algorithm can be used to construct minimal subsequential transducers.

Alternatively, the algorithm can be directly modified to cope with
final outputs. In this case, the equivalence relation $\fstequiv$
needs to be refined by requiring that two equivalent final states
must also have identical final outputs.

In some cases, the mapping $f: w\rightarrow o$ is not a function. For
example, a word in a pronunciation dictionary may have two or more
transcriptions. Such cases of bounded ambiguity can be handled by
another extension of the ST framework, namely {\em $p$-subsequential
transducers}, in which each final state is associated with up to $p$
final outputs \cite{Mohri:97}. The present algorithm can be extended
to this case by employing $p$ different end-of-string symbols
$\$_1\ldots \$_p$, as in the case of SSTs. This technique was used in
the application described in section~\ref{sec:eval}.


\subsection{Complexity and Optimisation}
\label{sec:compl}

For a dictionary of $m$ words, the main loop of the algorithm executes
$m$ times. The loops in procedures {\sc insert()} (including
the call to {\sc insert\_suffix()}) and {\sc remove\_duplicates()} 
are each executed $|w|$ times for each word $w$. Putting a state
on a register may be done in constant time when using a hash map. 

Compared to the FSA algorithm, the ST generalisation has one more
complexity component, namely the procedure {\sc push\_outputs()},
which is executed in each iteration of the loop in function {\sc
insert()}.  Each call to {\sc push\_outputs($q$)} involves
$\mathit{OutDegree(q)}$ operations.

In practical implementation, there is also some overhead stemming from
the use of more complex data structures (because of the need to
store transition outputs). This mainly affects the efficiency of
the register lookup and the {\sc insert()} procedure.

The algorithm can be optimised by reducing the number of times states
are registered/de-registered during the processing of the prefix of
$w$ in $T$ (main loop of {\sc insert()}). More precisely, the idea is
to deregister a state only if there is any residual output pushed down
the trie (i.e., the previous value of $\mathit{OutputSuffix}$ was
other than $\epsilon$). As a result, some states $q_1\ldots q_s$, $s <
k$, may stay registered after the call to {\sc insert()}. The loop in
{\sc remove\_duplicates()} must then check whether or not
$\delta(q_i,w_{i+1})=q_{i+1}$. If not, $q_{i+1}$ must have been
replaced by an equivalent state. In such a case, we must de-register
$q_i$ and check if there are equivalent states in the register. As
soon as one of the $q_i$'s is not replaced, there is no need to
perform this check for the remaining states $q_{i-1}\ldots q_1$.

This optimisation idea is used in the original algorithm. As for
STs, the speed-up achieved is moderate because {\sc
push\_outputs()} typically changes most of the outputs on the path
of $w$.

\section{Applications and Evaluation}
\label{sec:eval}

The new algorithm has been employed to construct pronunciation lexica
in the rVoice text-to-speech system.%
\footnote{{\scriptsize\tt
www.rhetorical.com/tts-en/technical/rvoice.html}}
In languages such as English, where the relation between orthography
and pronunciation is not straightforward, it is often advantageous to
store all known words in the dictionary, rather than rely on
letter-to-sound rules \cite{Fackrell:Skut:04}. The algorithm makes it
possible to store large amounts of data in such dictionaries without
affecting the efficiency and flexibility of the system: the resulting
representations are very compact, words can be looked up
deterministically in linear time, and user-defined entries can be
inserted into the dictionary at any time in any order (unlike in
Maurel and Mihov's approach). This last feature in particularly
important as rVoice users can control the behaviour of the system by
dynamically inserting their own entries into the dictionary at runtime.

The performance of the algorithm has been evaluated by constructing a
minimal ST for a pronunciation dictionary comprising the 50,000 most
frequent British surnames. The size and the construction time for the
ST were compared to the equivalent parameters for the sorted-data ST
algorithm \cite{Mihov:Maurel:01} and the unsorted-data FSA algorithm
\cite{Daciuk:ea:00}. The dictionary was not sorted, so there was an
extra sorting step in the case of Maurel and Mihov's algorithm
(sorting took less than 1 sec.  and is not included in the reported 
execution time). In the FSA, the phonetic transcriptions were encoded
as final emissions (i.e., the FSA encoded the language
$\{w^{(1)}o^{(1)},\ldots, w^{(m)} o^{(m)}\}$, each phonetic symbol
serving as an additional symbol of the input alphabet). The ST
encoding used a special end-of-string symbol \$ appended to each word
in order to make sure the resulting mapping was rational.%
\footnote{See section~\ref{sec:ext}. Transitions consuming \$
correspond to final emissions in a subsequential transducer. There
were 6,096 such transitions in the minimal ST.} The
results are shown in table~\ref{tab:exp}.

\begin{table}[h]
\begin{tabular}{l|l|l|l}
  & ST-unsorted  & ST-sorted & FSA \\ \hline
states & 22,211  & 22,211  & 161,592 \\ \hline
arcs & 67,129 &67,129 & 211,327 \\ \hline
time & 19 sec & 12 sec & 22 sec
\end{tabular}
\caption{Comparison of three construction methods (unsorted-data ST,
sorted-data ST and unsorted-data FSA) applied to a pronunciation
lexicon on a Pentium 4 1.7 GHz processor.}\label{tab:exp}
\end{table}
The comparison demonstrates that STs are superior to FSAs as an
encoding method for lexica annotated with rich respresentations. FSA
minimisation is obviously of little help if every (or almost every)
input is associated with a different annotation; almost no states are
merged in the part of the FSA encoding the $w^{(i)}$'s.%
\footnote{In contrast, the results reported by~Mihov and
Maurel~\shortcite{Mihov:Maurel:01} for a grammatical dictionary of
Bulgarian show little difference between the minimal FSA (47K states)
and the minimal ST (43K states). Clearly, this is due to the fact that
the number of grammatical classes is substantially smaller than the
number of word forms. } Since the FSA is much larger, construction
takes longer than in the ST case although the FSA algorithm is faster
on structures of equal size.

Not surprisingly, the sorted-data algorithm is faster than the
unsorted-data version, even including the actual sorting
time. However, its limited flexibility restricts its applicability,
leaving the new unsorted-data algorithm as the preferable option
in a range of applications.

\bibliography{../references/references.bib}
\end{document}